\documentclass{article}
\usepackage[a4paper,margin=3cm]{geometry}  % smaller margins
\usepackage{graphicx} % Required for inserting images
\usepackage{url}

\usepackage[dvipsnames]{xcolor}
\usepackage{microtype}
\usepackage{inconsolata}
\usepackage[T1]{fontenc}
\usepackage[utf8]{inputenc}
\usepackage{latexsym}
\usepackage{times}

\usepackage{enumitem}
\setlist[itemize]{leftmargin=1.2em,itemsep=0em,topsep=0.3em}

\usepackage{placeins}
\usepackage{float}
\usepackage{graphicx}
\usepackage{svg}
\usepackage{pgfplots}
\pgfplotsset{compat=1.18}
\usepackage{pgfplotstable}
\usepackage{graphicx} % for figures
\usepgfplotslibrary{groupplots}
\usepgfplotslibrary{colorbrewer}
\usepackage{wrapfig}
\usepackage{multicol}
\usepackage{multirow}
\usepackage{booktabs}

\usepackage{footmisc}        % provides \footref

\usepackage{xurl}

\usepackage{soul}

\newcommand{\grey}[1]{\textcolor{GreyColor}{#1}} 

\usepackage{natbib}
\usepackage{usebib}
\newbibfield{editor}

\setcitestyle{authoryear,round}
\usepackage{float}
\usepackage{adjustbox}
\usepackage{changepage}
\usepackage{lipsum}
\usepackage{tikz}
\usetikzlibrary{shapes, backgrounds}
\usepackage{amsmath}
\usepackage{hyperref}
\usepackage{amsmath}
\usepackage{xfrac}
\usepackage{titlesec}
\usepackage{tcolorbox}
\usepackage{ifthen}
\usepackage{xspace}
\usepackage{etoolbox}
\usepackage{listings}

\usepackage{caption}  % Include the caption package
\bibinput{custom}

\definecolor{RedColor}{HTML}{D9534F}
\definecolor{OrangeColor}{HTML}{F0AD4E}
\definecolor{BlueColor}{HTML}{5BB0DE}
\definecolor{GreenColor}{HTML}{5CB85C}
\definecolor{GreyColor}{HTML}{777777}
\definecolor{VioletColor}{HTML}{8E44AD}

\title{\textbf{Tendem:} A Hybrid AI+Human Platform}
\author{Toloka Team}
\date{November 2025}

\begin{document}
\maketitle

\begin{abstract}
Tendem is a hybrid system where AI handles structured, repeatable work and Human Experts step in when the models fail or to verify results. Each result undergoes a comprehensive quality review before delivery to the Client.

To assess Tendem’s performance, we conducted a series of in-house evaluations on 94 real-world tasks, comparing it with AI-only agents and human-only workflows carried out by Upwork freelancers. 

The results show that Tendem consistently delivers higher-quality outputs with faster turnaround times. At the same time, its operational costs remain comparable to human-only execution.

On third-party agentic benchmarks, Tendem's AI Agent (operating autonomously, without human involvement) performs near state-of-the-art on web browsing and tool-use tasks while demonstrating strong results in frontier domain knowledge and reasoning. %It is comparable to leading frontier models.

\end{abstract}

% ================================
\section{Introduction}\label{sec:intro}

We introduce Tendem\footnote{\url{https://tendem.ai}} -- a hybrid execution platform that joins AI execution with Human expertise that streamlines both professional and personal workflows. Tasks are completed through collaboration between the AI, which manages routine and fast-paced work, and a Human Expert who provides oversight, ensures contextual accuracy, and refines outputs.
If the AI reaches its limits, an expert continues the work to bring the task to completion. The platform’s architecture includes a multi-layered pipeline for task orchestration, routing, and quality assurance that integrates automated reasoning with human validation. \\

% We introduce Tendem\footnote{\url{https://tendem.ai}}, a hybrid human and AI execution platform that streamlines both professional and personal workflows. Tendem enables tasks to be completed through a collaborative process in which the AI handles routine and high-speed operations, while a Human Expert provides oversight, identifies gaps, refines outputs, and ensures contextual accuracy.

Unlike fully automated agentic workflows -- e.g., agents embedded in widely used assistants such as \emph{ChatGPT}\footnote{\url{https://chat.openai.com}} and \emph{Claude}\footnote{\url{https://www.anthropic.com/claude}} -- Tendem \emph{deliberately} retains a human in the loop for uncertainty- or impact-heavy steps. Purely autonomous agents tend to optimize for speed and cost but can struggle with under-specification, multi-document synthesis, or non-local quality criteria.

By contrast, human-only workflows, such as freelancer platforms, might offer traceable work history ratings and accountability but tend to be slower and more variable in cost. Tendem targets efficient points where it outperforms AI-only systems on quality for ambiguous or compliance-sensitive tasks while improving time-to-result and coordination cost relative to human-only execution.

% To evaluate the performance of hybrid human–AI collaboration, we need to benchmark the quality and efficiency of task execution across different operational modes. 

In this work, we present results from a comparative study involving fully automated pipeline, human-only workflow, and the Tendem hybrid model. We observe a substantial improvement in result quality and consistency: Tendem converts a much larger share of tasks into client-ready outcomes than either AI-only or human-only workflows. In addition, Tendem reduces operational cost relative to human-only execution, so higher quality does not come at the expense of price.

The evaluation comprises two components: an in-house set of 94 real-world tasks and a suite of public agentic benchmarks. We measure result quality (human evaluation), time, and price. 

Tendem achieves 74.5\% high-quality results across tasks and reduces median delivery time by 53\% versus the human-only baseline. We also evaluate Tendem's underlying AI agent (without human involvement) on external benchmarks. The agent is close to state-of-the-art on web browsing and real-world assistant tasks -- and remains close to leading models on hard knowledge. This provides a solid backbone for our Tendem Agent.

% ================================
\section{System Overview}\label{sec:overview}

% OLD:
% In this section, we present a high-level overview of the Tendem system -- its design principles, key roles, and the end-to-end pipeline from user input to the final response ready for delivery. We also discuss the mechanisms that prevent error accumulation while maintaining human focus where it matters, without introducing additional overhead.

% UPDATED:
Tendem’s architecture is built around the interaction between AI and Human Expertise. Each task moves through a structured flow that defines when the AI acts independently and when human judgment is required to ensure an efficient workflow and high-quality deliverables. Checks and verifications are built-in at every stage, so small errors do not accumulate, and human attention is directed to the points where it makes the greatest difference.

\subsection{Roles}

We focus on the following three roles when designing the system:

% \begin{itemize}
%     \item \textbf{Client / End User} -- defines a problem, desired outcomes and acceptance criteria; values quality, price and time; expects transparency and privacy controls.
    
%     \item \textbf{Tendem's AI Agent} -- handles clarification, produces a structured plan and executes and auto-verify the steps. Produces traceable artifacts (files, citations, logs) and auto-checks. Designed with `hybrid' use case in mind -- efficient and streamless communication with an expert. 
    
%     \item \textbf{Human Expert} -- performs targeted intervention at critical points, corrects, judges, and polishes outputs; expects AI agent to be helpful and reliable. The Human Expert is a highly skilled professional with domain experience who ensures that the result meets all requirements.
% \end{itemize}

\subsubsection*{Client / End User}

The \textbf{Client} begins by outlining the problem and describing desired outcomes, sharing any background material or data needed to work effectively. Clients expect results that meet their standards, within a limited time-frame, and at a consistent cost. But just as importantly, they expect a clear view of how their data is handled and protected at every stage.

\subsubsection*{Automated Component: Tendem's AI Agent and Tools}

\textbf{Tendem's AI Agent} executes the bulk of routine and tool-heavy work via a plan–act–observe–verify loop with explicit step gates for verification \citep{yao2023reactsynergizingreasoningacting,weng2023largelanguagemodelsbetter,manakul2023selfcheckgptzeroresourceblackboxhallucination}. Tooling includes \emph{web browsing}, \emph{file I/O} for common office formats, a safe and isolated \emph{Python} runtime for analytics and charting, and an isolated secure runtime environment \emph{bash/CLI}. All actions run with minimal permissions and are recorded to make every step traceable and repeatable.

Quality assurance tools run continuously: spec conformance, unit/total reconciliation, citation matching, and lightweight self-consistency checks. When the system encounters uncertainty -- whether from conflicting sources, failed checks, or a step that carries too much risk -- it escalates to a Human Expert.

\subsubsection*{Human Expert}

A \textbf{Human Expert} is a professional who adds the kind of reasoning and contextual awareness that models cannot yet provide \citep{Huang_2025,Ji_2023,bender-koller-2020-climbing}. They guide the system through uncertain steps, verifying, and refining the final result for real-world use. Experts are admitted through a series of tests and exams to to verify their expertise and ensure they undergo training and continuous QA.

In production workflow, Experts intervene at step gates (plan audit, plan step check, draft refinement, etc.) and perform the final offline QA pass as part of the layered-QA. The performance is tracked using QA and rework rates.

\subsection{End-to-End Workflow}

Tendem workflow consists of the following interconnected steps:

\begin{itemize}
    \item \textbf{Client request}: the Client outlines what they want to achieve in plain language, adding any files or references the system will need to understand the task.
    
    \item \textbf{Clarify and formalize}: the AI Agent inspects files and requirements and asks targeted questions.
    
    \item \textbf{Plan with gated steps}: the AI Agent decomposes the task into steps, gating critical and high-risk ones under Human Expert supervision.
    
    \item \textbf{Routing and matching}: the system identifies a suitable expert based on the required skill set and generates a time estimate.
    
    \item \textbf{Hybrid execution}: the AI agent carries out the work under human supervision, with the Human Expert stepping in whenever the output needs adjustment or further development.
    
    \item \textbf{Online QA}: the system performs lightweight automated checks of both AI Agent and human edits, updating the plan and iterating further steps if needed.
    
    \item \textbf{Offline QA}: after task execution, the system conducts automated multi-step verification against the customer's requirements and attached materials, escalating uncertain cases to a human QA expert.
    
    \item \textbf{Finalization}: the system delivers the verified result to the Client or returns it for rework if material issues remain.
\end{itemize}

\subsection{Target Advantages}

Having a multi-component workflow helps us achieve several key advantages:

\begin{enumerate}[label=(\arabic*)]
    % \item Consistently high result quality through hybrid collaboration between agent and expert, reinforced by layered QA processes.
    
    % \item Increased expert productivity, as the agent autonomously handles routine subtasks, accelerates overall task completion, and increases attention to details.
    
    % \item Improved Client experience -- the structured flow, quick and targeted clarifications, and transparent checkpoints make interaction with the system more intuitive, reliable, and efficient.

    \item \textbf{Intelligent Hybrid Execution:} Task decomposition and dynamic routing (AI for speed, Human Expert for judgment, and a verification loop between them) reduce coordination overhead and ensure every deliverable benefits from both machine efficiency and expert human judgment.
    
    \item \textbf{Multi-Layer Quality Assurance:} Each deliverable is first tested through automated systems that flag potential faults or signs of hallucination. From there, a human reviewer traces how the output was produced, checking its logic and alignment with the source material and requirements until it meets the standard expected for delivery.

    \item \textbf{Expert productivity uplift.} Routine subtasks are automated, allowing Human Experts to concentrate on high-impact decisions and polish, which shortens total cycle time and increases attention to detail.

    \item \textbf{Improved Client experience.} The system designed to have clear structured flow, targeted clarifications and visible progress, to make working with it feels straightforward and efficient.
\end{enumerate}

% \subsubsection*{Intended Use and Scope}
% \textbf{Best fit:} \red{designed for all, domain-agnostic; but now we focus on the general and simple coding} routine business tasks (research, structured writing, data cleaning, marketing operations, basic analytics, scripting, simple software engineering). \\
% \textbf{Out-of-scope:} regulated advice and high-stakes decisions without human review (e.g. medical, legal); tasks that require expert-level knowledge (e.g. scientific, math). Limitations on LLM / Human Experts with special expertiese. But still the system designed to be  domain-agnostic. Just not all experts yet `hired'.

% ================================
\section{Evaluation}\label{sec:eval}

To quantify price–quality trade-offs, we evaluate Tendem on in-house collected, economically valuable tasks that mirror real freelance platforms. We compare Tendem against AI agents and a freelance-marketplace worker baseline (\autoref{sec:eval_internal}). This measures client-perceived outcomes on valuable real-life tasks.

To ensure transparent comparison, we also evaluate Tendem's underlying AI agent (without human intervention) on public benchmarks against existing models and AI-only agents (\autoref{sec:eval_external}). We use it as a proxy for assessing the strength of the backbone AI agent and its impact on the hybrid system’s overall efficiency and performance.

% ================================
\subsection{In-house benchmark with real-world tasks}\label{sec:eval_internal}

Public benchmarks probe core skills but underrepresent real-life settings that depend on a Client's personal preferences (ambiguous briefs, complex multi-stage tasks, input and files as deliverables).

To better assess system performance on \emph{real-world, economically valuable remote-work projects} that reflect the complexity and practical constraints of professional task execution, we introduce a new in-house benchmark. It approximates real freelance-platform task distributions to mirror the conditions of professional work.

We assume that customers evaluate performance and successful task completion across three independent dimensions: \textbf{result quality}, \textbf{execution time}, and \textbf{price}.

\subsubsection*{Metrics}
To compare Tendem’s end-to-end performance against other AI agents and a freelance-platform baseline, we measure the following for each task:

\begin{enumerate}[leftmargin=1.2em]
    \item \textbf{Result Quality}: each output is evaluated on a three-point scale (Bad, Mediocre, Good), with additional Decline label. We assess quality across three fine-grained quality criteria and overall result quality.
    
    \begin{itemize}
        \item \textbf{Accuracy} -- factual and numerical correctness, with no fabrications, sources, or outdated information; the presence of a reference does not compensate for an inaccuracy, it should be relevant.
        
        \item \textbf{Completeness} -- coverage of all required items, constraints, and deliverables in the Client's task so the result is self-contained and usable without follow-up, no items are missed, no part of the processing omitted.
        
        \item \textbf{Style \& Formatting} -- clarity, structure, and presentation quality, including readable organization (headings, lists, tables), consistent units and notation, correct grammar/spelling, appropriate tone, and correct output format.
        
        \item \textbf{Overall Quality} -- end-to-end fitness for purpose: the answer is aligned with the user’s intent and ready to deliver given the three criteria above (assigned independently, to weigh contextual factors and practical usefulness).
    \end{itemize}

    Each criterion is graded using the following scale:
    
    \begin{itemize}
        \item \textbf{Good} — full criteria satisfaction, client-ready; at most nitpicks that do not require changes.
        \item \textbf{Mediocre} — requires some edits to satisfy the criteria. 
        \item \textbf{Bad} — materially incomplete; not suitable without substantial rework.
        \item \textbf{Decline} — task refused or not attempted due to safety filters or domain coverage limits.
    \end{itemize}
    
  \item \textbf{Time} (latency): measured as connect time (if any) from task posting to active work, execution time to final result, and total time as the sum of both.
  
  \item \textbf{Price}: the effective cost of each task, calculated as the sum of human labor and backbone AI usage, recorded in USD.
\end{enumerate}

\subsubsection*{Evaluated Systems}

For evaluations, we selected commonly used AI-only system and human-only system. The following were evaluated:

\begin{enumerate}
  \item \textbf{ChatGPT Agent (OpenAI)\footnote{\url{https://openai.com/index/introducing-chatgpt-agent/}}.} A multi-step agentic assistant embedded in ChatGPT with actions/tool use and browsing. Evaluated via the ChatGPT UI with default settings.
  
  % \item \textbf{Manus\footnote{\url{https://manus.im/blog/manus-1.5-release}}.} A commercial agentic assistant with multi-step planning, integrated web browsing, and tool use. We evaluated Manus via its public UI using a `Pro' subscription with default settings.
  
  \item \textbf{Upwork\footnote{\url{https://www.upwork.com/}\label{fn:upwork}}.} Human freelancers hired through the Upwork marketplace who follow principles described below (best-match, price, job-success filter). Freelancers are hired independently for each task. We did not instruct them to use or not to use AI. % No mandated AI assistance. 
  
  % \item \textbf{Tendem's underlying AI agent\footnote{\url{https://tendem.ai/}}.} The Tendem's AI Agent runs fully autonomously without human intervention. We report two budget modes -- \emph{performance} (larger thinking budgets, larger models under the hood, deeper verification) and \emph{efficiency} (tighter budgets, faster models).

  \item \textbf{Tendem\footnote{\url{https://tendem.ai/}}.} The production version of Tendem. The detailed system overview can be found in \autoref{sec:overview}.
\end{enumerate}

\subsubsection*{Dataset Design and Collection}

Building a representative, real-life benchmark of Client tasks involved the following structured process:

\begin{itemize}[leftmargin=1.2em]
    \item \textbf{Use-case sampling.} Common tasks categories were selected from Upwork\footref{fn:upwork}, Fiverr\footnote{\url{https://www.fiverr.com/}}, and Freelancer\footnote{\url{https://www.freelancer.com/}} to approximate the distribution of paid knowledge work.
    
    \item \textbf{Task authoring.} Based on the sampled categories, we commissioned practitioners to author new tasks with clear acceptance criteria and attached assets (e.g., spreadsheets, PDFs, links).
    
    \item \textbf{PII and compliance.} Tasks were created to be \textbf{PII-free} and avoid regulated domains (e.g., medical/legal advice). 
\end{itemize}

The final benchmark area/category distribution is presented in Table~\ref{table:internal_distribution}.
\begin{table}[!ht]
    \centering
    \small
    \begin{tabular}{llc}
    \toprule
    \textbf{Area} & \textbf{Category} & \textbf{Count} \\
    \midrule
    \multirow{2}{*}{Sales} 
        & Collect Business Contact Data & 14 \\
        & Complete Missing Fields (enrichment) & 6 \\
    \cmidrule{2-3}
    \multicolumn{2}{r}{\textbf{Sales Total}} & \textbf{20} \\
    \midrule
    \multirow{7}{*}{Operations}
        & Build Multi-step Automation Workflows & 1 \\
        & Collect Data & 10 \\
        & Convert Formats & 2 \\
        & Retrieve PDF / Document / Report Content & 3 \\
        & Schedule \& Manage Appointments \& Calls & 3 \\
        & Structure Raw Data into Schema & 6 \\
        & Validate Contact Info & 3 \\
    \cmidrule{2-3}
    \multicolumn{2}{r}{\textbf{Operations Total}} & \textbf{28} \\
    \midrule
    \multirow{4}{*}{Marketing}
        & Collect Business Contact Data & 4 \\
        & Create Content & 7 \\
        & Market \& Competitive Research Reports & 10 \\
        & Proofread, analyse and correct content & 3 \\
    \cmidrule{2-3}
    \multicolumn{2}{r}{\textbf{Marketing Total}} & \textbf{24} \\
    \midrule
    \multirow{4}{*}{Analysis}
        & Customer / User Interviews or Feedback Collection & 1 \\
        & Generate Performance Dashboards \& Summaries & 8 \\
        & Market \& Competitive Research Reports & 8 \\
        & Run Exploratory Data Analysis & 5 \\
    \cmidrule{2-3}
    \multicolumn{2}{r}{\textbf{Analyst Total}} & \textbf{22} \\
    \midrule
    \multicolumn{2}{l}{\textbf{Grand Total}} & \textbf{94} \\
    \bottomrule
    \end{tabular}
    \caption{In-house benchmark area and category distribution.}
    \label{table:internal_distribution}
\end{table}

\subsubsection*{Task submissions protocol}

All third-party systems were used via their public UIs -- hand-pasting task text -- to reflect the end-user experience and capture real connection latency. The highest subscription tier is used where available to ensure the highest rate limits and best backbone models. We captured timestamps when each task was created, when human work started, and when the final result was delivered.

\subsubsection*{Freelancer Selection at Upwork}

Because freelance platforms offer experts with varying rates and skill levels, the following principles were used to guide selection to ensure fair comparison.

Tasks were posted and proposals collected; selection prioritized previous human experience, price, job success percentage, using information from the Upwork profile of applicants. We preferred the least expensive candidate with over 80\% job success or promising newcomers. The performer was selected without price negotiation -- bids were evaluated and the lowest acceptable offer was accepted as-is.

\subsubsection*{Labeling Protocol}

We recruited independent human QA experts from our professional network and onboarded them, requiring passage of an entrance exam. Raters received concise instructions on how to apply the evaluation criteria (Accuracy, Completeness, Style \& Formatting, Overall) and example-driven guidance. To reduce bias, \emph{the systems were anonymized} and evaluated in the same interface. The evaluators saw only the input task, attached assets, and the produced outputs. All experts received a competitive hourly payment.

For each task, a single Expert assigned labels (\emph{Good, Mediocre, Bad} for all 4 quality scales, and brief notes justifying the decision. We record a \emph{Decline} when a system refused to complete the task. For this experiment, we used single-rater judgments per task; but a large fraction of items received spot checks to ensure high quality labeling. 

We did not employ or validate LLM-as-a-judge \citep{gu2025surveyllmasajudge} for scoring, so all results rely on human evaluation only (yet it can be a good follow-up experiment).

\subsubsection*{Results}

% TODO: add std +-
\pgfplotstableread[col sep=comma]{
System,Bad,Mediocre,Good,Decline,Acc_Good,Comp_Good,Style_Good,Price_Avg,Price_Med,Price_Avg,Price_Med,Time_Total_Med
Tendem,8.5,16.0,74.5,1.1,74.5,81.9,70.2,69.2,32.0,69.2,32.0,16.42
Upwork,21.3,25.5,53.2,0.0,63.8,59.6,59.6,48.0,50.0,48.0,50.0,34.97
% Manus,22.3,38.3,39.4,0.0,57.4,51.1,57.4,5.2,2.6,5.2,2.6,0.23
% ChatGPT Agent,44.7,26.6,25.5,3.2,35.1,41.5,43.6,0,0,0,0,0.13  OLD
ChatGPT Agent,36.2,19.1,40.4,4.3,48.9,48.9,51.1,0,0,0,0,0.13
}\alldata

% ===== Quality distribution (Bad/Mediocre/Good) =====
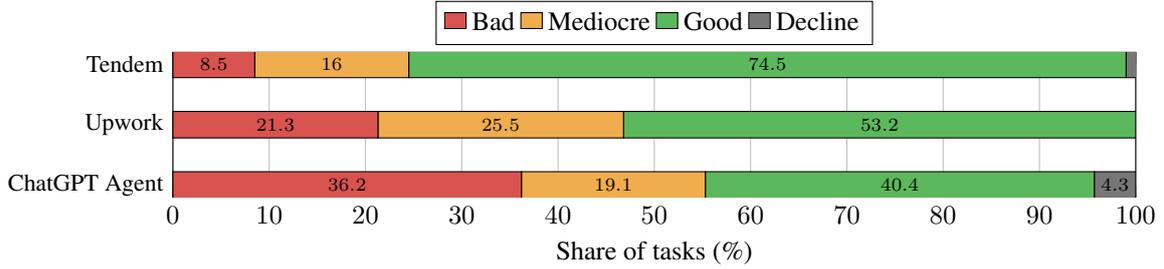
\begin{figure}[H]
    \centering
    \begin{tikzpicture}
    \begin{axis}[
      xbar stacked,
      width=0.95\linewidth, height=3.5cm,
      xmin=0, xmax=100,
      xmajorgrids,
      symbolic y coords={ChatGPT Agent,Upwork,Tendem},
      ytick=data,
      yticklabel style={font=\small},
      xlabel={Share of tasks (\%)},
      legend style={at={(0.5,1.05)},anchor=south,legend columns=4},
      nodes near coords, nodes near coords align={horizontal},
      every node near coord/.append style={font=\scriptsize},
      nodes near coords align={center},
    ]
    \addplot[fill=RedColor]    table[x=Bad,    y=System]{\alldata};
    \addplot[fill=OrangeColor]    table[x=Mediocre,y=System]{\alldata};
    \addplot[fill=GreenColor]   table[x=Good,   y=System]{\alldata};
    \addplot[fill=GreyColor]   table[x=Decline,   y=System]{\alldata};
    \legend{Bad,Mediocre,Good,Decline}
    \end{axis}
    \end{tikzpicture}
    \caption{The fraction of tasks rated Bad/Mediocre/Good for each system on in-house benchmark. Each system evaluated by a human without system name indication. Decline label means safety filters of the system declined the task.}
    \label{chart:internal_quality}
\end{figure}

\paragraph{Systems Evaluation.} 
Figure~\ref{chart:internal_quality} summarizes the distribution of Bad/Mediocre/Good outcomes for each evaluated system. Tendem attains 74.5\% Good, outperforming Upwork by +21.3 pp (53.2\%) and ChatGPT Agent by +34.1 pp (40.4\%). 

Tendem also exhibits fewer weak outcomes: 8.5\% Bad vs 21.3\% (Upwork), and 36.2\% (ChatGPT Agent); showing itself as a more stable and reliable performer. 

Overall, \textbf{Tendem’s Good rate is significantly higher than all baselines} (Upwork and ChatGPT Agent) on our in-house evaluation. A one-sided \( z \)-test comparing the share of `Good' results between Tendem and the closest baseline Upwork confirms the improvement is statistically significant (\( \text{p-value} = 0.0012 \)).

\begin{table}[H]
    \centering
    \small
    \begin{adjustbox}{width=1.0\linewidth}
    \begin{tabular}{l|cccc|ccc}
        \toprule
        \multirow{2}{*}{\textbf{System}}  & \multicolumn{4}{c}{\textbf{Overall}} & \textbf{Accuracy} & \textbf{Completeness} & \textbf{Style} \\
        \cmidrule(lr){2-5}
         & \footnotesize{(\% Good)} & \footnotesize{(\% Mediocre)} & \footnotesize{(\% Bad)} & \footnotesize{(\% Decline)} & \footnotesize{(\% Good)} & \footnotesize{(\% Good)} & \footnotesize{(\% Good)} \\
        \midrule
        Tendem        & \textbf{74.5} \grey{$\pm$ 4.5} & 16.0 \grey{$\pm$ 3.8} & 8.5 \grey{$\pm$ 2.9} & 1.1 \grey{$\pm$ 1.0} & \textbf{74.5} \grey{$\pm$ 4.5} & \textbf{81.9} \grey{$\pm$ 4.0} & \textbf{70.2} \grey{$\pm$ 4.7} \\
        Upwork        & 53.2 \grey{$\pm$ 5.1} & 25.5 \grey{$\pm$ 4.5} & 21.3 \grey{$\pm$ 4.2} & 0.0 \grey{$\pm$ 0.0} & 63.8 \grey{$\pm$ 5.0} & 59.6 \grey{$\pm$ 5.1} & 59.6 \grey{$\pm$ 5.1} \\
        % Manus         & 39.4 \grey{$\pm$ 5.1} & 38.3 \grey{$\pm$ 5.0} & 22.3 \grey{$\pm$ 4.3} & 0.0 \grey{$\pm$ 0.0} & 57.4 \grey{$\pm$ 5.1} & 51.1 \grey{$\pm$ 5.2} & 57.4 \grey{$\pm$ 5.1} \\
        ChatGPT Agent & 40.4 \grey{$\pm$ 5.1} & 19.1 \grey{$\pm$ 4.1} & 36.2 \grey{$\pm$ 5.0} & 4.3 \grey{$\pm$ 2.1} & 48.9 \grey{$\pm$ 5.2} & 48.9 \grey{$\pm$ 5.1} & 51.1 \grey{$\pm$ 5.2} \\
        \bottomrule
    \end{tabular}
    \end{adjustbox}
    \caption{The fraction of tasks rated Bad, Mediocre, and Good for each system on the in-house benchmark for overall and three quality criteria. Decline label means system declined the task due to safety filters or lack of expertise for the task. Each system was evaluated by a human without system name indication. Bold font indicates best result in column.}
    \label{table:internal_quality}
\end{table}

\paragraph{Analysis of Quality Criteria.}
In Table~\ref{table:internal_quality} we go beyond overall task evaluation and analyze performance across three quality criteria: 

\begin{itemize}[leftmargin=1.25em]
  \item \emph{Accuracy.} Tendem improves Accuracy vs. Upwork by +10.7 pp (74.5\% vs 63.8\%), indicating that designed hybrid workflow and layered QA strengthens factual grounding and numerical correctness.
  
  \item \emph{Completeness.} Largest gap: +22.3 pp vs Upwork (81.9\% vs 59.6\%). One of the reasons might be a workflow with step-gates that reduce omissions and enforce acceptance criteria.
  
  \item \emph{Style \& Formatting.} Tendem gains +10.6 pp vs Upwork (70.2\% vs 59.6\%) and +19.1 pp vs ChatGPT Agent (70.2\% vs 51.1\%), yielding cleaner, client-ready presentation.
\end{itemize}

\textbf{Error patterns.} AI-only systems exhibit (i) omissions on long specs, (ii) shallow synthesis beyond model context limits, (iii) fabricated references \citep{Ji_2023,Huang_2025}. Human-only outputs show uneven quality along all criteria. Tendem substantially cuts omission/fabrication errors via escalation at step gates with high uncertainty.

\begin{table}[H]
    \centering
    \small
    \begin{adjustbox}{width=1.0\linewidth}
    \begin{tabular}{l|cc|cccccc}
    \toprule
    \multirow{3}{*}{\textbf{System}}
        & \multicolumn{2}{c|}{\multirow{2}{*}{\textbf{Price} (USD)}}
        & \multicolumn{6}{c}{\textbf{Time} (hours)} \\
    \cmidrule(lr){4-9}
        & \multicolumn{2}{c|}{} 
        & \multicolumn{2}{c}{\textbf{Connection}}
        & \multicolumn{2}{c}{\textbf{Execution}}
        & \multicolumn{2}{c}{\textbf{Total}} \\
    \cmidrule(lr){4-5}\cmidrule(lr){6-7}\cmidrule(lr){8-9}
        & Average & Median 
        & Average & Median & Average & Median & Average & Median \\
    \midrule
    Tendem        
        & 69.2 \grey{$\pm$ 9.3} 
        & 32.0 \grey{$\pm$ 5.3} 
        & 4.8 \grey{$\pm$ 5.7} & 2.6 \grey{$\pm$ 0.6} 
        & 20.9 \grey{$\pm$ 24.3} & 10.5 \grey{$\pm$ 3.8} 
        & 25.7 \grey{$\pm$ 24.3} & 16.4 \grey{$\pm$ 2.9} \\
    Upwork        
        & 48.0 \grey{$\pm$ 3.3} 
        & 50.0 \grey{$\pm$ 3.8} 
        & 14.5 \grey{$\pm$ 27.4} & 4.6 \grey{$\pm$ 0.5} 
        & 38.3 \grey{$\pm$ 54.3} & 26.8 \grey{$\pm$ 3.9} 
        & 52.7 \grey{$\pm$ 58.8} & 35.0 \grey{$\pm$ 6.8} \\
    % Manus         
    %     & 5.2 \grey{$\pm$ 0.7} 
    %     & 2.6 \grey{$\pm$ 0.5} 
    %     & -- & -- 
    %     & 22.6 \grey{$\pm$ 122.7} & 0.2 \grey{$\pm$ 0.03} 
    %     & 22.6 \grey{$\pm$ 122.7} & 0.2 \grey{$\pm$ 0.03} \\
    ChatGPT Agent 
        & -- & -- 
        & 0 & 0 
        & 0.2 \grey{$\pm$ 0.2} & 0.1 \grey{$\pm$ 0.01} 
        & 0.2 \grey{$\pm$ 0.2} & 0.1 \grey{$\pm$ 0.01} \\
    \bottomrule
    \end{tabular}
    \end{adjustbox}
    \caption{Price and timing metrics with standard deviation shown as \grey{$\pm$ std}. Prices are in USD; times are in hours. For median values std is calculated using bootstrap.  
    ChatGPT Agent is subscription-based service, hence no direct price.}
    \label{table:internal_price_time}
\end{table}

\paragraph{Price and Speed.}
Table~\ref{table:internal_price_time} provides execution time (split by \emph{connection} and effective \emph{execution} time) and price analysis per task.

Tendem’s median price is 32.0 USD versus Upwork’s 50.0 USD (-36\%); its average price is higher at 69.2 USD versus 48.0 USD for Upwork, reflecting a right-tailed cost distribution with a few high-touch cases.

When it comes to time, Tendem reduces median total time from 35.0 h (Upwork) to 16.5 h (-53\%), with faster connection (2.6 h versus 4.6 h, -42\%) and execution (10.5 h versus 26.8 h, -61\%). ChatGPT Agent exhibits very low latency and low price as expected, but -- as shown in Table~\ref{table:internal_quality} -- delivers lower quality on this dataset.

Overall, \textbf{Tendem delivers results significantly faster than the human-only baseline and at a lower median price, with higher quality}. The higher average price reflects a small number of complex, escalated tasks that required additional expert involvement.

% ===== Pareto: Quality vs time with optimal corner =====
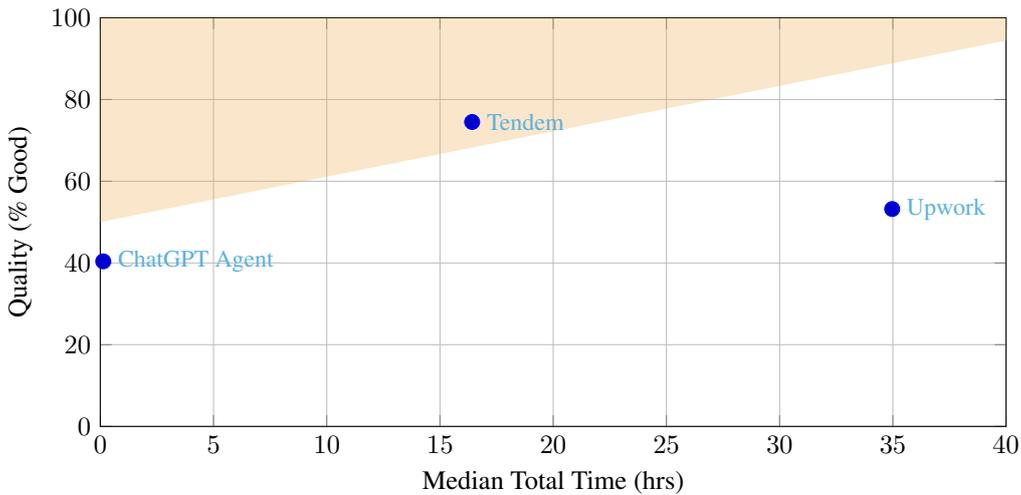
\begin{figure}[H]
    \centering
    \begin{tikzpicture}
    \begin{axis}[
      width=0.90\linewidth, height=7cm,
      xlabel={Median Total Time (hrs)},
      ylabel={Quality (\% Good)},
      xmin=0, xmax=40, ymin=0, ymax=100,
      xmajorgrids, ymajorgrids,
      % For label text from the table:
      point meta=explicit symbolic
    ]
      % --- Light-grey optimal corner (upper-left).
      \fill[OrangeColor, opacity=0.3] (axis cs:0,100) -- (axis cs:45,100) -- (axis cs:0,50) -- cycle;

      % --- Points with labels sourced from the table (no duplication)
      \addplot+[
        only marks,
        mark=*,
        mark size=2.8pt,
        nodes near coords,
        nodes near coords style={font=\small, color=BlueColor, anchor=west, xshift=2pt}
      ] table[
        x=Time_Total_Med, y=Good, meta=System
      ] {\alldata};

    \end{axis}
    \end{tikzpicture}
    \caption{Evaluated system quality vs median Total Time. Light-orange corner highlights the desirable region (lower time, higher quality).}
    \label{chart:internal_pareto}
\end{figure}

Figure~\ref{chart:internal_pareto} shows the trade-off between quality versus total time. Tendem occupies a favorable region (higher quality at a similar or lower time than Upwork), showing best quality per hour spent. Despite lower quality, AI-only system lies on the sub-efficient frontier, while they can solve some tasks fast and cheap -- many outputs are low quality and fail to meet requirements.

\paragraph{Takeaways.} 

\begin{itemize}
    \item \textbf{Quality:} Tendem improves overall Good rate by +21.3 pp vs Upwork (74.5\% vs 53.2\%), with the largest gain in \emph{Completeness} (+22.3 pp).
    
    \item \textbf{Speed:} Tendem cuts median total time by an impressive 53\% (16.5 h vs 35.0 h), driven by faster both connection and execution times.
    
    \item \textbf{Cost:} Typical (median) price is 36\% lower than Upwork (32.0 vs 50.0 USD), while average cost is higher due to a small number of escalated, high-touch tasks.
    
    \item \textbf{Positioning:} For ambiguous, multi-document, or spec-heavy tasks, hybrid step-gates reduce omissions/fabrication, \textbf{closing the last-mile gap where AI-only systems remain brittle}.
\end{itemize}

% ================================
\subsection{External Benchmarks (Tendem's underlying AI agent)}\label{sec:eval_external}

Additionally, the underlying AI agent -- operating autonomously without human involvement -- was evaluated to assess the strength of the automation layer and its performance relative to other AI agents.

\subsubsection*{Benchmarks}

We include benchmarks that (i) reflect \emph{real-world tasks} over toy puzzles, (ii) are \emph{hard}, not saturated at $\approx$100\%, and (iii) match our target scope of browsing/tool use and factual reasoning (iv) are not \emph{environment-constrained} (no predefined tool implementations).

\begin{itemize}
  \item \textbf{BrowseComp} \citep{browsecomp2025}: deep, multi-hop web browsing for verifiable answers.
  \item \textbf{HLE (Humanity’s Last Exam)} \citep{hle2025}: hard multi-domain knowledge and reasoning.
  \item \textbf{GAIA} \citep{gaia2311}: artificially created assistant tasks with tool-use/browsing.
  % \item \textbf{GDPval} \citep{gdpval2025}: evaluation on \emph{economically valuable, real-world tasks} across 1320 tasks with offline evaluation on OpenAi's side.
\end{itemize}

\subsubsection*{Setup and policy}
All Tendem runs in this section use the autonomous backbone AI agent from the Tendem system, evaluated independently.

Tendem's AI agent was run on all benchmark tasks using the specified configuration, with official benchmark evaluation scripts applied to the results.

Evaluation follows official metrics: exact match score for BrowseComp GAIA and HLE \citep{browsecomp2025,gaia2311,hle2025}. For contextual baselines, we use official system cards/blogposts values as-is; setups might differ across sources. We chose it to ensure a fair comparison with Tendem.

\subsubsection*{Results}

Table \ref{table:external_results} compares Tendem's AI Agent against our baselines.

Tendem's AI Agent is competitive on browsing/tool use and performs strongly relative to established baselines, yet leaves room for improvement.

\begin{table}[!ht]
    \centering
    % \begin{adjustbox}{width=1.0\linewidth}
    \begin{tabular}{lccc}
    \toprule
    \textbf{System} & \textbf{BrowseComp} & \textbf{HLE} & \textbf{GAIA} \\
    \midrule
    Tendem's AI agent \tiny{without human involvement}  & \textbf{71.0} & 39.0 & \textbf{78.2} \\
    % Tendem's AI agent \tiny{without human involvement, efficiency}   & 62.5 & 28.0 & 74.5 \\
    ChatGPT Agent                        & 68.9 & 41.6 & --    \\
    ChatGPT Deep Research                & 51.5 & 26.6 & 67.4 \\
    GPT-5 pro \tiny{with tools}          & --    & \textbf{42.0} & --    \\
    GPT-5 high \tiny{with tools}         & 54.9 & 35.2 & --    \\
    % GPT-5 high \tiny{no tools}           & --    & 24.8 & --    \\
    o3 high \tiny{with tools}            & 49.7 & 24.3 & --    \\
    Perplexity Deep Research             & --    & 21.1 & --    \\
    Manus                             & --    & --    & 73.4 \\
    Flowith                              & --    & --    & 78.1 \\
    HF Open Deep Research \tiny{GPT-5 medium} & --    &  --     & 62.8 \\
    HAL Generalist Agent  \tiny{Claude Sonnet 4.5} & --    &  --    & 72.6 \\
    \bottomrule
    \end{tabular}
    % \end{adjustbox}
    \caption{Evaluation of Tendem and other AI systems on agentic benchmarks. All values are percentages. Non-Tendem numbers are vendor-reported; setups might differ across sources.}
    \label{table:external_results}
\end{table}

Tendem's AI Agent is \textbf{competitive on web browsing/tool-use} (state-of-the-art tier on BrowseComp; top on GAIA) and \textbf{close to leading models on hard knowledge} (HLE), yet still has a room for improvement. Results support the hybrid agent: high-quality agent to automate routine/tool-heavy steps and escalate judgment where models remain brittle.

\subsection{Evaluation limitations}\label{sec:external_eval_limitations}

The in-house benchmark reflects the distribution of typical freelance tasks but does not fully represent the wider market nor include specialist work such as senior-level programming, medical, or legal-advice tasks.

Also, the style of the system outputs might affect evaluations; however we try to mitigate this bias by removing any reference to the evaluated system, using sub-scale rubrics to calibrate raters, and allowing a expert to report uncertainty where relevant.

Cross-benchmark comparability is imperfect: BrowseComp uses live web; HLE disallows browsing by design yet many models report score with web tools; GAIA mixes tool use and retrieval; HAL reports are cost-aware but configuration-sensitive~\citep{browsecomp2025,gaia2311,hle2025}.

% ================================
\section{Safety, Intended Use, and Limitations}
\label{sec:safety}

% \textbf{Intended use.} Routine business tasks that rely on tool use and produce verifiable outputs, with human review available when judgment is required..\\
\textbf{Out-of-scope.} Regulated or expert-level advice (yet we are expanding domains coverage); high-stakes decisions.

\paragraph{Primary risks and mitigations.}
\begin{itemize}
  \item \textbf{Hallucinations / Fabrication.} Step gates and source-required checks; offline QA blocks handoff when confidence is low.
  
  \item \textbf{Specification drift.} Acceptance criteria captured during clarification; enforced as tests in online QA.
  
  \item \textbf{Privacy.} All experts working with Tendem operate in a secure environment (VM) and follow strong privacy protocols. Additionally, we perform regular production-use inspections, enforce access controls, audit experts.

\end{itemize}

\section{Release and Reproducibility}\label{sec:release}

We release our in-house collected benchmark with economically valuable tasks. We publish full input text, output and produced files for each of the systems -- on GitHub \url{https://github.com/toloka/tendem-evaluation}.

% ================================
\section{Conclusion}\label{sec:conclusions}

Tendem is a hybrid agent that combines automated planning and tool use with human step gates and layered quality assurance. On a 94-task in-house benchmark that mirrors economically valuable freelance tasks, Tendem delivers higher quality and faster turnaround than both human-only and AI-only approaches, achieving 74.5\% Good outcomes and a 53\% reduction in median total time compared to freelance workers.

The observed quality gains arise from the full hybrid stack rather than any single component. High-quality backbone AI agent executes routine and tool-heavy steps and Human Experts and QA specialists provide targeted review and updates.

Tendem's AI agent evaluation shows that the underlying agent operating autonomously is competitive on browsing and real-world assistant tasks, and remains close to leading models on hard knowledge, supporting the design choice to automate routine, tool-heavy steps while escalating judgment-heavy decisions to Human Experts.

We will continue improving both Agentic and Hybrid components by expanding tool integrations, strengthening online/offline QA signals, refining the routing and escalation pipeline, and broadening the expert pool from general to more specialized domains.

% ================================
\subsection*{Core Contributors}

Konstantin Chernyshev, 
Ekaterina Artemova, 
Viacheslav Zhukov, 
Maksim Nerush, 
Mariia Fedorova, 
Iryna Repik, 
Olga Shapovalova, 
Aleksey Sukhorosov, 
Vladimir Dobrovolskii, 
Natalia Mikhailova, 
Sergei Tilga

\FloatBarrier
\bibliography{custom}

\appendix

% % ================================
% \section{Case Studies}

% \subsubsection*{CS1: TBA}

% \red{TBA}

% \subsubsection*{CS2: TBA}

% \red{TBA}

\end{document}